\documentclass[10pt,twocolumn,letterpaper]{article}
\usepackage[utf8]{inputenc}
\usepackage{cvpr}
\usepackage{times}
\usepackage{epsfig}
\usepackage{graphicx}
\usepackage{amsmath}
\usepackage{stmaryrd}
\usepackage{amssymb}
\usepackage{multirow}
\usepackage{verbatim}
\usepackage{color}
\usepackage{float}
\usepackage{enumitem}
\usepackage{booktabs}
\usepackage{subcaption}
\usepackage{makecell}
\usepackage{soul}
\usepackage{mathtools}
\usepackage{epigraph}
\usepackage[pagebackref=true,breaklinks=true,letterpaper=true,colorlinks,bookmarks=false]{hyperref}
\hypersetup{
	plainpages=false,
	colorlinks=true,              %
	linkcolor=blue,               %
	anchorcolor=blue,             %
	citecolor=blue,               %
	filecolor=blue,               %
	urlcolor=blue,                %
	pdfview=FitH,                 %
	pdfstartview=FitH,            %
	pdfpagelayout=SinglePage      %
}

\cvprfinalcopy % *** Uncomment this line for the final submission
\usepackage{pifont}
\newcommand{\cmark}{\ding{51}}%
\newcommand{\xmark}{\ding{55}}%
\def\cvprPaperID{2151} % *** Enter the CVPR Paper ID here

\newcommand{\rYC}{YR10}
\newcommand{\rMSRVTT}{MR10}
\newcommand{\rCrossTask}{CTR}
\newcommand{\rCOIN}{FA}

\definecolor{mygray}{gray}{0.6}
\newcommand{\ResNetFifty}{R50}
\newcommand{\KinSeven}{K700}
\newcommand{\KinFour}{K400}
\newcommand{\KinSix}{K600}
\newcommand{\HowToM}{HTM}
\newcommand{\CrossTask}{CT}
\newcommand{\ImageNet}{ImNet}
\newcommand{\tablestyle}[2]{\setlength{\tabcolsep}{#1}\renewcommand{\arraystretch}{#2}\centering\footnotesize}
\ifcvprfinal\fi
\begin{document}

%%%%%%%%% TITLE

\title{End-to-End Learning of Visual Representations \\ from Uncurated Instructional Videos }
\author{
	Antoine Miech\textsuperscript{1}\footnotemark[1]
	\quad\quad\quad
	Jean-Baptiste Alayrac\textsuperscript{2}\footnotemark[1]
	\quad\quad\quad
	Lucas Smaira\textsuperscript{2}
	\\
		\quad\quad
	Ivan Laptev\textsuperscript{1}
	\quad\quad\quad\quad\quad
	Josef Sivic\textsuperscript{1,3}
	\quad\quad\quad
	Andrew Zisserman\textsuperscript{2,4}
	\\
	\small{$^1$ENS/Inria \quad $^2$\normalsize{DeepMind} \small \quad $^3$CIIRC CTU \quad $^4$VGG Oxford}
	\\
	\small{\texttt{antoine.miech@inria.fr,jalayrac@google.com}}
	\\
	\small{\url{https://www.di.ens.fr/willow/research/mil-nce/}}
}

\pagestyle{plain}

\twocolumn[{%
\renewcommand\twocolumn[1][]{#1}%
\vspace{-0.5cm}
\maketitle

\begin{center}
\includegraphics[width=\linewidth]{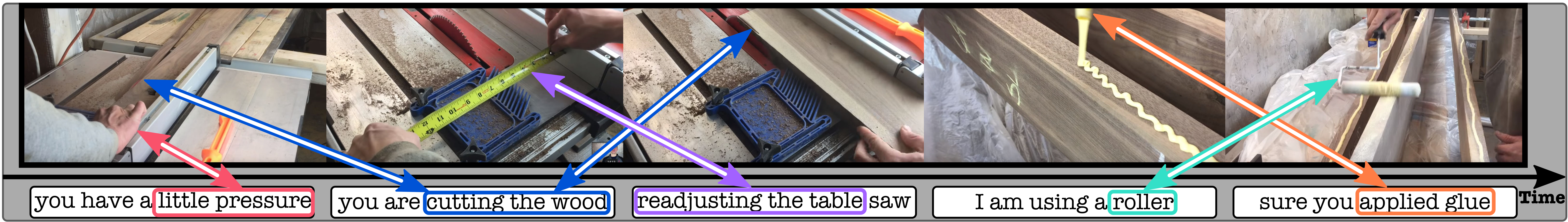} 
\captionof{figure}{	\small 
We describe an efficient approach to learn visual representations from misaligned and noisy narrations (bottom) automatically extracted from instructional videos (top). Our video representations are learnt \emph{from scratch without relying on any manually annotated visual dataset} yet outperform all self-supervised and many fully-supervised methods on several video recognition benchmarks.   
}
\label{fig:teaser}
\end{center}%
}]

%%%%%%%%% ABSTRACT
\begin{abstract}
Annotating videos is cumbersome, expensive and not scalable.
Yet, many strong video models still rely on manually annotated data.
With the recent introduction of the HowTo100M dataset, narrated videos now offer the possibility of learning video representations without manual supervision.
In this work we propose a new learning approach, 
\emph{MIL-NCE}, capable of addressing misalignments inherent in narrated videos.
With this approach we are able to learn strong video representations from scratch, without the need for any manual annotation. 
We evaluate our representations on a wide range of four downstream tasks over eight datasets: action recognition (HMDB-51, UCF-101, Kinetics-700), text-to-video retrieval (YouCook2, MSR-VTT), action localization (YouTube-8M Segments, CrossTask) and action segmentation (COIN).
Our method outperforms all published self-supervised approaches for these tasks as well as several fully supervised baselines. 
Our joint text-video pretrained model is publicly available at: \url{https://www.di.ens.fr/willow/research/mil-nce/}.
\end{abstract}

%%%%%%%%% BODY TEXT
\section{Introduction}

\epigraph{What we see changes what we know. \\ What we know changes what we see. \hspace{1.5em} \textit{Jean Piaget}}{}
% Possible nice quote if we want to go that way?
Vision and language play an important role in the way humans learn to associate visual entities to abstract concepts and vice versa.
This has also become the \emph{de facto} way to successfully train computer vision models.
Indeed, from \emph{classification} where images are categorized based on a fixed list of words to the recent \emph{captioning} tasks where images or videos are annotated with rich language descriptions, this interplay is one of the driving forces behind recent progress in the field.
However, one of the main limitations of this approach is that it requires manually annotating large collections of visual data.

\renewcommand{\thefootnote}{\fnsymbol{footnote}}
\footnotetext[1]{Equal contribution.}
\renewcommand*{\thefootnote}{\arabic{footnote}}
\footnotetext[1]{D\'{e}partement d'informatique de l'ENS, \'{E}cole normale sup\'{e}rieure, CNRS, PSL Research University, 75005 Paris, France.}
\footnotetext[3]{Czech Institute of Informatics, Robotics and Cybernetics at the Czech Technical University in Prague.}
\footnotetext[4]{VGG, Dept.\  of Engineering Science, University of Oxford}

Manual annotation is both cumbersome and expensive.
Moreover, for videos, which are the main focus of this work, annotation is also even more challenging due to the  ambiguities of choosing the right vocabulary of actions and annotating action intervals in video.
This significantly limits the scale at which fully supervised video datasets can be obtained and hence slows down the quest to improve visual representations.
Recent work has proposed a promising alternative to this fully supervised approach:  leveraging narrated videos that are readily available at scale on the web.

Of particular interest, the recent HowTo100M dataset~\cite{miech19howto100m} contains more than 100 million pairs of video clips and associated narrations.
It was automatically collected by querying YouTube for instructional videos.
Such videos usually depict someone explaining orally how to perform a complex human activity, \eg preparing a particular meal or repairing a car.
Our objective in this paper is to learn strong video representations using \emph{only} this narrated material.

End-to-end learning from instructional videos is a highly challenging task.
Indeed, these videos are made in general with the goal of maximizing the number of views, and with no specific intention to provide a training signal for machine learning algorithms.
This means that the supervision present in the narration is only weak and noisy.
Among typical sources of noise, the principal one by far is the weak {\em alignment} between the video and the language: although for the most part the spoken words correlate with what is happening in these videos, this alignment is far from perfect.
People might talk about something \emph{before} actually demonstrating it, but they might also \emph{omit} to talk about something that is happening because it is clear enough visually.
Conversely they could only mention an action without showing it in the case where the step is not essential or trivial to convey with language alone.
This is without even considering the {\em irrelevant} information given throughout the video (\eg jokes or credits) as well as the general difficulty of working with spoken language obtained from potentially \emph{erroneous} speech recognition algorithm as opposed to written text.

In this work, we propose a bespoke training loss, dubbed \emph{MIL-NCE} as it inherits from Multiple Instance Learning (MIL) \emph{and} Noise Contrastive Estimation (NCE).
Our method is capable of addressing visually misaligned narrations from uncurated instructional videos as illustrated in Figure~\ref{fig:teaser}.
Equipped with this novel training scheme and a simple joint video and text embedding model, we show that we can successfully train video representations \emph{from scratch} directly from pixels on the HowTo100M~\cite{miech19howto100m} dataset.
To demonstrate the quality of the learnt representations, we employ an extensive set of evaluation benchmarks on a wide variety of video understanding tasks: action recognition (HMDB-51, UCF-101, Kinetics-700), text-to-video retrieval (YouCook2, MSR-VTT), action localization (YouTube-8M Segments, CrossTask) and action segmentation (COIN).
Notably, our learnt video representations outperform fully supervised baselines trained on Kinetics or ImageNet for several of the tasks.
We also show improvements over other self-supervised approaches on HMDB51 and UCF101 even without fine-tuning the learnt representations.
Finally, by leveraging the joint video and text representations, our off-the-shelf trained model also reaches state-of-the-art results on YouCook2 and CrossTask, without any training on the target datasets.

\noindent
\textbf{Contributions.}
The contributions of this work are threefold.
\textbf{(i)}~We propose a method to learn a joint text video embedding in an end-to-end fashion from unlabelled, uncurated narrated videos using the recently introduced HowTo100M~\cite{miech19howto100m} dataset. 
In particular, we introduce a specific loss, dubbed \emph{MIL-NCE} for Multiple Instance Learning Noise Contrastive Estimation, that enables the learning to cope with the highly misaligned narration descriptions.
\textbf{(ii)}~We provide a thorough ablation study to quantitatively assess the importance of the different design choices of the approach.
\textbf{(iii)}~Finally, we demonstrate that the representations thus obtained are competitive with their strongly supervised counterparts on four downstream tasks over eight video datasets.

\begin{figure*}[t]
      \centering
		\begin{subfigure}[t]{.63\linewidth}
			\centering
		\includegraphics[width=\linewidth]{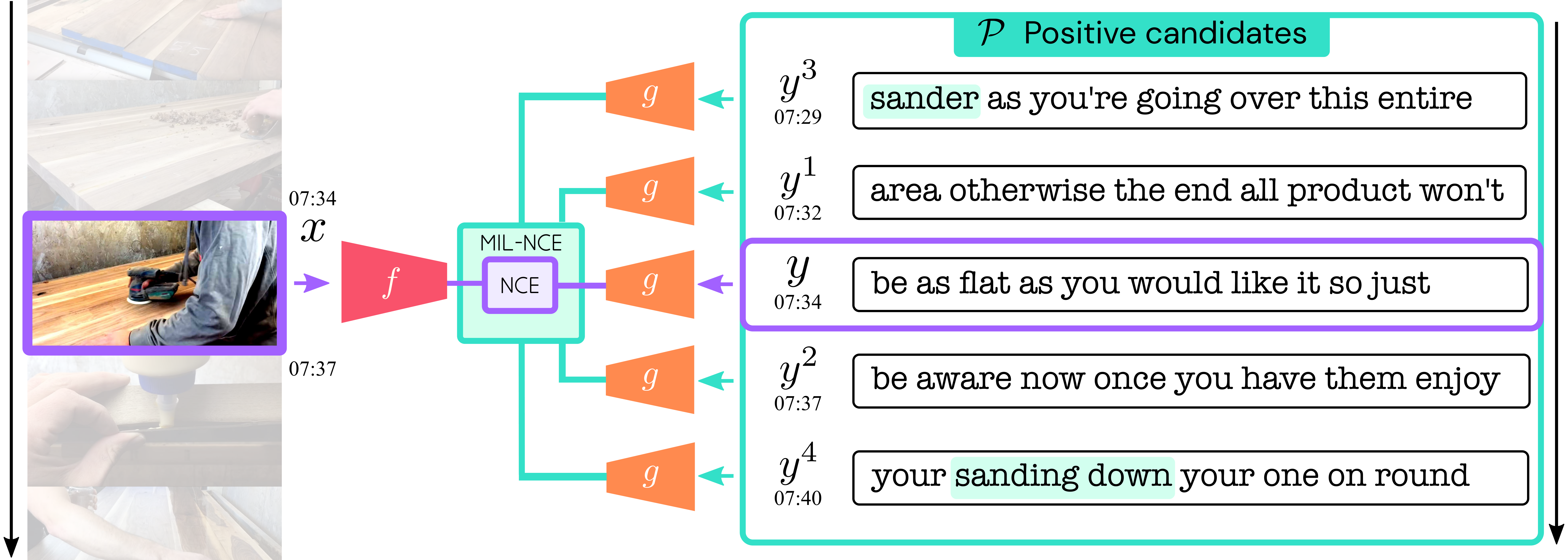}
			\caption{\label{fig:mil_vizu} \small \textbf{Examples of positive candidates}}
		\end{subfigure}
			\hfill
			\begin{subfigure}[t]{.345\linewidth}
		\centering
		\includegraphics[width=\linewidth]{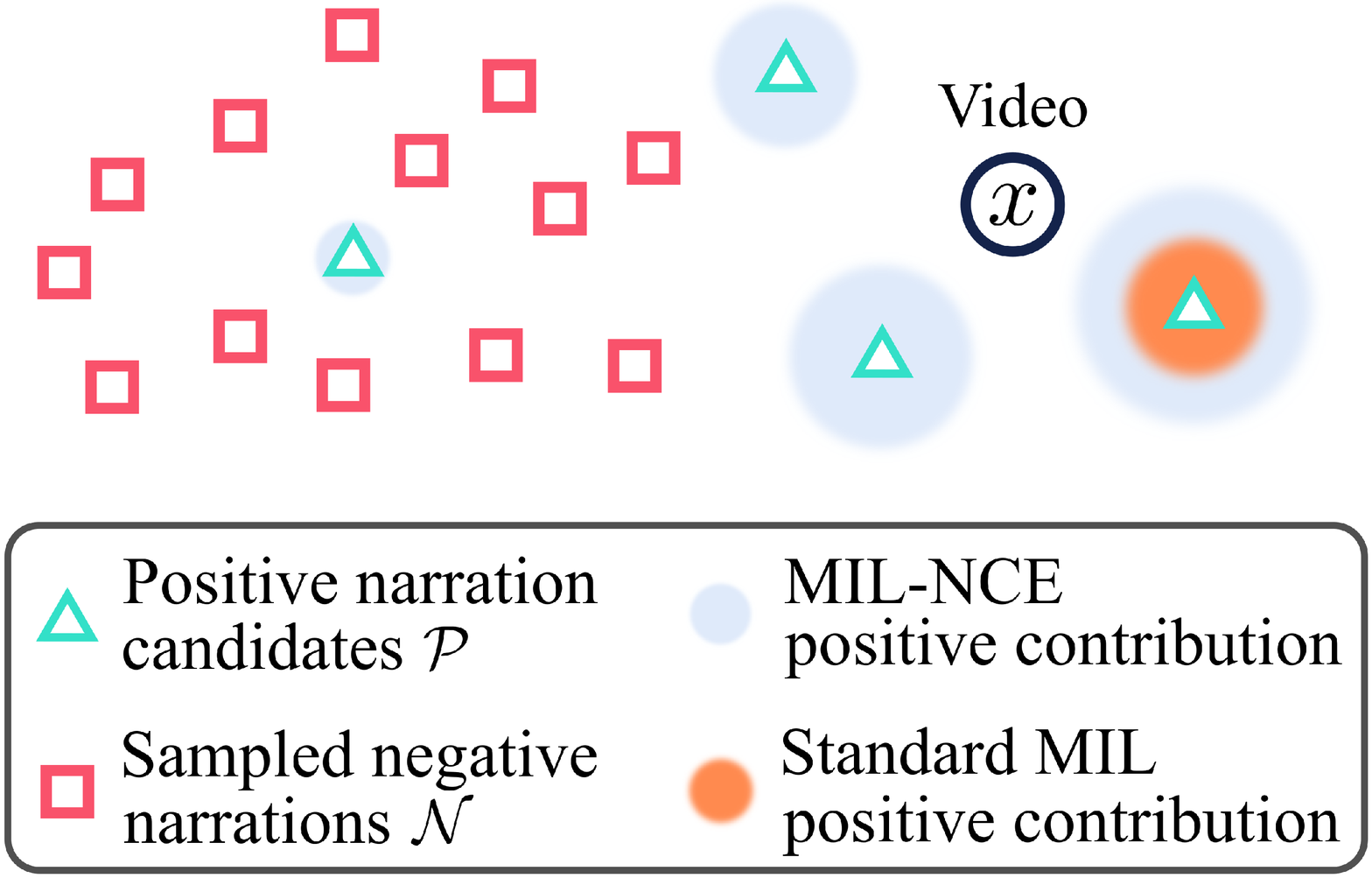}
		\caption{\small \textbf{Illustration of \emph{MIL-NCE}} \label{fig:mil_nce}}
	\end{subfigure}
	\caption{\label{fig:model} \small \textbf{Left.}
Our MIL-NCE makes it possible to consider a set of  multiple positive candidate
pairs $\{(x, y), (x, y^1), \dots, (x, y^4)\}$ while the standard
NCE approach would only consider the single $(x, y)$ training pair
and miss the visually grounded object description
\texttt{sander} from pair $(x, y^3)$ or the action description \texttt{sanding down} from $(x, y^4)$.
\textbf{Right.} 
Given a video $x$ and an associated set of positive narration candidates $\mathcal{P}$ (green triangles) that may or may not be correct, our \emph{MIL-NCE} selects \emph{multiple} correct positives (large blue areas) while downweighting incorrect positives (smaller blue areas) based on a discriminative ratio against
negatives $\mathcal{N}$ (red squares). In contrast, traditional MIL considers only one positive (orange circle) while discarding the rest. 
}
\end{figure*}

\section{Related work}
%We review below, works that are closely related to ours.

\noindent
\textbf{Learning visual representations from unlabeled videos.}
As labeling videos is cumbersome, expensive and not scalable, a significant number of prior works have
studied the task of learning visual representations from unlabeled videos.
Currently, the most effective approach is to collect a large amount of data from social media and use the available metadata as supervision~\cite{youtube8m,ghadiyaram2019large}.
However, this metadata is often in the form of keywords or tags, rather than (spoken) natural language considered in this work. In addition, the meta data is often platform dependent and rarely publicly available.
Self-supervised approaches do not suffer from these issues as the idea is to define a supervised proxy task using labels directly generated from videos.
Some of these tasks include: temporal ordering of video clips or frames~\cite{fernando2017self, lee2017unsupervised,misra2016shuffle,xu2019self},
predicting geometric transformations~\cite{jing2018self}, maximizing the mutual information of multiple views~\cite{tian2019contrastive}, predicting motion and appearance~\cite{wang2019self}, predicting the future, the past or a portion of masked input in the feature space~\cite{han2019video, sun2019contrastive, vondrick2016anticipating}, colorizing videos~\cite{vondrick2018tracking}, predicting 3D geometry from synthetic data~\cite{gan2018geometry}, predicting the audio in a feature space~\cite{arandjelovic2018objects,korbar2018cooperative} or tasks leveraging temporal cycle consistency~\cite{dwibedi2019temporal,wang2015unsupervised}.
In this work our proxy task is supervised by the automatic speech recognition (ASR) applied to narrated instructional videos.
The nature of this supervision has the potential to also provide semantic information~\cite{miech19howto100m,sanabria18how2}, which is often missing in works that only exploit pixel-wise cues.
Moreover, most of the top performing prior works only study their method on curated video datasets (\eg Kinetics~\cite{carreira2017quovadis}) where labels have been removed.
However, this is not truly learning from unlabeled data as these videos have been carefully selected and verified to belong to classes of interests.
Caron \etal~\cite{caron2019unsupervised} further explain the performance gap between training on such curated data versus uncurated ones, truly available at scale.
Instead, our approach focuses on the learning of representations \emph{only from uncurated videos}.

\noindent
\textbf{Vision, speech and language.}
A common alternative to training visual models using manually defined sets of labels is to exploit semantic supervision from natural language or speech.
Numerous prior works~\cite{chowdhury2018webly,dong19dual,gong14multi,gong14improving,klein15associating,miech18learning,
mithun2018learning,pan16jointly,plummer2017enhancing,xu2015jointly,
wang2018learning,wang2016learning,wray2019fine,wu2017sampling} have used image / video description datasets~\cite{lin14coco, plummer2015flickr30k, rohrbach17movie, xu16msrvtt, youcook2} to learn an embedding space where visual and textual data are close only if they are semantically similar.
These methods either rely on manually annotated image / video description datasets, or leverage representations already pre-trained on manually labelled datasets (\eg ImageNet~\cite{imagenet} or Kinetics~\cite{carreira2017quovadis}). 
In contrast, in this work \textit{no manually annotated visual data is involved at any stage of our approach.}
To avoid labeling visual data, several approaches have leveraged audio transcripts obtained from narrated videos using automatic speech recognition (ASR) as a way to supervise video models for object detection~\cite{amrani2019toward, chen17discover, moriya2019grounding}, captioning~\cite{hessel2019case,sun2019videobert}, classification~\cite{alayrac16unsupervised,kuehne2019mining, malmaud15what,yu14instructional}, summarization~\cite{palaskar2019multimodal} or retrieval~\cite{miech19howto100m} using large-scale narrated video datasets such as How2~\cite{sanabria18how2} or HowTo100M~\cite{miech19howto100m}.
Others~\cite{boggust2019grounding, harwath18jointly} have investigated learning from narrated videos by directly using the raw speech waveform instead of generating transcriptions.
Most related to us is the work of Miech \etal~\cite{miech19howto100m} who trained a joint video and text embedding from uncurated instructional videos~\cite{miech19howto100m}.
However, as opposed to our work, they do not model any misalignment issue encountered when training on such videos and 
rely on visual representations pretrained on Kinetics-400 and ImageNet.
Building on this work, Sun~\etal~\cite{sun2019contrastive} have used a contrastive bidirectional transformer (CBT) to learn long term contextual video representations from instructional videos. 
All these works use a visual representation pre-trained on either Kinetics or ImageNet when training on such narrated videos.
In contrast, the key innovation of our work is that we demonstrate learning a generic video representation as well as a joint video-text embedding {\em from scratch}, without pre-training on manually annotated video or image datasets.

\noindent
\textbf{Multiple instance learning for video understanding.}
Multiple instance learning methods have been employed in many weakly-supervised video understanding problems including:
person recognition in movies using scripts~\cite{bojanowski13finding,miech17learning,parkhi15it}, anomaly detection~\cite{sultani2018real},
weakly supervised action classification~\cite{leung2011handling,shapovalova12similarity} and localization~\cite{cheron2018flexible,duchenne09automatic,weinzaepfel16towards}, co-reference resolution of characters in TV series~\cite{ramanathan14linking} or object tracking~\cite{babenko2009visual}.
These methods often rely on some form of max-pooling (\ie MIL-SVM~\cite{andrews2003support}) or discriminative clustering (\ie DIFFRAC~\cite{bach07diffrac}) to resolve the label ambiguities, and have used mostly linear (or shallow) models. % and did not show strong improvements when trained with recent deep learning models.
In this work, we present MIL-NCE, a new approach marrying the noise contrastive estimation (NCE) framework~\cite{gutmann2010noise} with multiple instance learning~\cite{dietterich1997solving}. We show that MIL-NCE is well-suited to learn deep visual representations from scratch using the weak and noisy training signals available in uncurated instructional videos.

%%%%%%%%%%%%%%%%%%%%%%%%%%%%%
\section{Leveraging Uncurated Instructional Videos}
%%%%%%%%%%%%%%%%%%%%%%%%%%%%%

This section describes the proposed approach to train joint video and text embeddings from unlabeled narrated videos in an end-to-end fashion.
To start with, we are given $n$ pairs of video clips and associated narrations.
In practice, a pair is composed of a short 3.2 seconds video clip (32 frames at 10 FPS) together with a small number of words (not exceeding 16) that correspond to what the person is saying in the video.
For example, someone might be \emph{sanding wood} while mentioning the action ``\texttt{sanding down}'' or the object ``\texttt{sander}'' as illustrated in Figure~\ref{fig:mil_vizu}.
Given this input, our goal is to {\em learn a joint embedding space}  where similarity between the narration and video embedding is high when the text and visual content are semantically similar and low otherwise,
and we wish to learn this starting from raw pixels in the video and text descriptions.
As illustrated in Figure~\ref{fig:teaser}, this is a very challenging
problem due to the often severely misaligned visual descriptions.

In this work, we address this issue by introducing the \emph{MIL-NCE} objective:
%, called \emph{MIL-NCE} due to its link with Multiple Instance Learning~\cite{dietterich1997solving} and Noise Contrastive Estimation~\cite{gutmann2010noise}:
\begin{equation}
\label{eq:objective}
\max_{f,g}\sum_{i=1}^n\log\hspace*{-1mm}\left(\frac{\sum\limits_{(x,y)\in\mathcal{P}_i}e^{ f(x)^\top g(y)}}{\hspace*{-1mm}\sum\limits_{(x,y)\in\mathcal{P}_i}\hspace*{-2mm}e^{ f(x)^\top g(y)}+\hspace*{-5mm}\sum\limits_{(x',y')\sim\mathcal{N}_i}\hspace*{-4mm}e^{f(x')^\top g(y')}}\right)\hspace*{-1mm}
\end{equation}
where $x$ represents a video clip and $y$ a narration.
$f$ and $g$ are the two embedding functions that respectively operate over video and text.
Given a specific sample (indexed by $i$), we construct $\mathcal{P}_i$ to be a valid set of \emph{positive} video/narration candidate pairs (see Figure~\ref{fig:model}) while $\mathcal{N}_i$ conversely refers to an associated set of negative video/narration pairs.
%Given a specific sample $i$-th, $\mathcal{P}_i$ stands for specific set of video/narration pairs that we consider as potential \emph{positives} (see Figure~\ref{fig:model}) while $\mathcal{N}_i$ conversely refers to a set of negative video/narration pairs.
This objective function implies \emph{maximizing the ratio} of the sum of the positive candidate scores from $\mathcal{P}_i$ to the sum of the scores of \emph{all} negatives sampled from $\mathcal{N}_i$, where the score is measured by the exponentiated dot product of the corresponding video and language embeddings, $f(x)$ and $g(y)$.
%This offers a \emph{strong} learning signal and contrasts with other standard metric learning losses such as the~\emph{triplet loss}~\cite{chechik2010large} that only consider individual pairwise comparisons between positives and candidates.

%This objective function can be interpreted as \emph{maximizing the total contribution} of the  scores of the positive candidates from $\mathcal{P}_i$ over the total contribution of the negatives sampled from $\mathcal{N}_i$, where the score is measured by the exponentiated dot product of the corresponding video and language embeddings, $f(x)$ and $g(y)$.

In the following, we describe more precisely the motivation behind the \emph{MIL-NCE} objective~\eqref{eq:objective}.
First, Section~\ref{sec:model} introduces the chosen probabilistic model for joint text and video embedding.
Given that model, Section~\ref{sec:losses} details the choice behind the training objective~\eqref{eq:objective} explaining how it is specifically adapted to handle the misalignment noise inherent in narrated videos in comparison with existing approaches.

\subsection{A simple joint probabilistic model}
\label{sec:model}
In the following, $x\in\mathcal{X}$ stands for a video clip and $y\in\mathcal{Y}$ for a narration.
 % Because the goal is to learn directly from pixels, $\mathcal{X}$ is of the form $\mathbb{R}^{T\times W\times H \times 3}$, where $T$ is the number of frames, $W$ and $H$ are the width and height of the frames, and there are 3 standard RGB channels.
% Recall we focus on short clips of 3.2 seconds (32 frames at 10 FPS).
% az suggest leaving out the following def
% $\mathcal{Y}$ takes the form of $\llbracket 0, M\rrbracket^L$, where $M$ is the number of words in our dictionary (with $0$ being a special null token) and $L$ is the maximum number of words considered per narration.
% We use narrations provided in the HowTo100M dataset~\cite{miech19howto100m} that are rather short with $L$ never exceeding 16 words (see Figures~\ref{fig:noise} and~\ref{fig:mil_vizu} for examples).
Given a set of $n$ pairs of video clips and associated narrations $\{(x_i,y_i)\}_{i=1}^ n\in (\mathcal{X}\times\mathcal{Y})^n$  sampled from the joint data distribution $P(\mathcal{X}\times\mathcal{Y})$,
our goal is to learn a joint embedding space where semantically related videos and texts are close and far away otherwise.

Formally, we learn two parametrized mappings: $f:\mathcal{X}\to\mathbb{R}^d$ maps a video clip $x$ into a $d$-dimensional vector $f(x)\in\mathbb{R}^d$,  and $g:\mathcal{Y}\to\mathbb{R}^d$ maps a narration $y$ into the same $d$-dimensional vector space, $g(y)\in\mathbb{R}^d$.
We assume that we can estimate up to a constant factor the joint probability of a pair of video and narration $(x,y)$ by exponentiating the dot product of the two embeddings:
\begin{equation}
\label{eq:prob_model}
p(x,y; f, g) \propto e^{f(x)^\top g(y)}.
\end{equation}
In this work, $f$ takes the form of a CNN that runs over a fixed-length clip.
For $g$, we consider simple sentence based models that transform a set of words into a single vector.
Note,  for simplicity and with a slight abuse of notation, we refer to $f$ (or $g$) as both a function \emph{and} the parameters that define it.
Also, we will refer to~\eqref{eq:prob_model} as simply $p(x,y)$, \ie we keep the dependence in $f$ and $g$ \emph{implicit} for the clarity of simpler equations.
More details about the exact architecture of the models are provided in Section~\ref{sec:exp}.
% Next, we explain how to train such models from \emph{scratch} using only unlabeled narrated videos.

\subsection{Learning from uncurated data: MIL-NCE}
\label{sec:losses}

Recall that our goal is to learn a joint video and text representation only from \emph{uncurated} narrated videos.
In this section, we start by detailing why this is a highly challenging endeavor due to misalignments present in that data.
Next, we explain how the introduced \emph{MIL-NCE} objective~\eqref{eq:objective} enables to learn despite that noise.
Finally, we contrast our proposed approach to similar works in the self-supervised domain.

\noindent
\textbf{Misalignment in narrated videos.}
In \cite{miech19howto100m}, the authors estimate that around 50\% of clip-narration pairs from the HowTo100M dataset are not aligned.
In fact, people are likely to describe an event after or before performing it in the video as illustrated in Figure~\ref{fig:teaser}.
%Figure~\ref{fig:teaser} illustrates four common types of noise to which we can attribute this issue:
%\textbf{(i)} people often talk about things that are \emph{irrelevant} to the visual scene  (\eg credits, jokes ...),
%\textbf{(ii)} the Automatic Speech Recognition (ASR) from which transcripts are obtained can \emph{fail},
%\textbf{(iii)} people may \emph{omit} to describe what they are doing or conversely to do what they are saying,
%and finally \textbf{(iv)} we face a \emph{temporal alignment} problem as people often talk about something before or after demonstrating it, {\em and} the description can be spread over several clips.
This temporal misalignment makes it more challenging to learn video representations than with manually annotated and aligned labels.

\noindent
\textbf{How to learn despite noisy supervision?}
To address the aforementioned issues, we propose to consider multiple options for matching a video and a narration instead of \emph{only} comparing a single video $x$ with a single narration $y$ as done in~\cite{miech19howto100m}.
Let's consider the example illustrated in Figure~\ref{fig:mil_vizu}.
Given a clip $x$, $K$ narrations $\{y_k\}_{k=1}^K$ that happen close in time within the same video can be considered as positive candidates.
By doing so, the chance that spoken words correlate with what is happening in the video increases.
In that case, we would like to match \emph{at least one} of the narrations $\{y_k\}_{k=1}^K$ with video $x$.
Given the probabilistic model~\eqref{eq:prob_model}, a natural way to express this is by computing the joint probability of $x$ happening with any of the $y_k$.
Because we can make the assumption that $y_k$'s are mutually exclusive (\ie $(x,y_i)\neq(x,y_j), \forall i\neq j$), this can be expressed mathematically by~\eqref{eq:prob_model} as follows:
\begin{equation}
\label{eq:core}
p(\cup_k \{(x, y_k)\})=\sum_k p(x,y_k) \propto \sum_k e^{ f(x)^\top g(y_k)}.
\end{equation}
This is a MIL extension, which, as opposed to MIL-SVM, does not explicitly select a single positive sample per bag at training.
More generally, and symmetrically, the case where several video clips are candidates for a given narration can also be envisioned.
Hence, for generality, we assume that instead of having a single pair $(x, y)$, we have a set of candidate positive pairs $\mathcal{P}={(x^k,y^k)}_{k=1}^K$, and we can simply repurpose~\eqref{eq:core} as $p(\mathcal{P}) \propto \sum_{(x,y)\in\mathcal{P}} e^{f(x)^\top g(y)}$.
We denote by $\{\mathcal{P}_i\}_{i=1}^n$ the training set of candidate positive pairs deduced from the original training set $\{(x_i,y_i)\}_{i=1}^n$.
With this extension, we have the tools to address misalignments.
Details about how to construct $\mathcal{P}_i$ are given in Section~\ref{subsec:impl}

\noindent
\textbf{How to train this model?}
We wish to learn a video representation based on the previously described probabilistic model $p(\mathcal{P})$.
However, this is challenging as one cannot directly apply standard \emph{generative} techniques such as maximum likelihood due to the intractability of computing the normalization constant over all possible pairs of videos and narrations.
Instead, we rely on a \emph{discriminative} technique, namely the noise-contrastive estimation (\emph{NCE}) approach~\cite{gutmann2010noise,jozefowicz2016exploring}, that has recently been shown to be effective in the context of feature learning~\cite{henaff2019data,oord2018representation}.
The core idea is to directly optimize the \emph{unnormalized} probabilistic model~\eqref{eq:core} to discriminate between data obtained from the true joint distribution $P(\mathcal{X}\times\mathcal{Y})$ and some artificially generated noise data, also called ``negatives''. 
In this work, we use the softmax version of NCE~\cite{jozefowicz2016exploring}:
\begin{equation}
\max_{f,g}\sum_{i=1}^n\log\left(\frac{e^{ f(x_i)^\top g(y_i)}}{e^{ f(x_i)^\top g(y_i)}+\sum\limits_{(x',y')\sim\mathcal{N}_i}\hspace*{-4mm}e^{ f(x')^\top g(y')}}\right)
\end{equation}
and replacing the probability of a single positive match, $e^{ f(x_i)^\top g(y_i)}$, with
our MIL like extension, $\sum_{(x,y)\in\mathcal{P}_i} e^{f(x)^\top g(y)}$, gives our proposed
\emph{MIL-NCE} training objective~\eqref{eq:objective}.
Given this, we can simply estimate the parameters of our model by maximizing the objective~\eqref{eq:objective}, where $\mathcal{N}_i$ is a specific set of negatives for the $i$-th sample.
%In Section~\ref{sec:exp}, details about what are good choices for $\mathcal{N}_i$ are discussed.
Next, we discuss how our approach differs from prior work.

\noindent
\textbf{NCE objectives for self-supervised learning.}
NCE has recently been successfully applied to self-supervised learning.
In particular, CPC~\cite{henaff2019data,oord2018representation} introduces the \emph{InfoNCE} loss to enforce the model to maximize the conditional probability of some targets (\eg the bottom part of the image) conditioned on some context (\eg the top part of the image).
Differently from CPC, which creates an \emph{asymmetric} set of negatives by fixing the context and only sampling negative targets, we instead use NCE to model the symmetric joint probability between text and video~\eqref{eq:prob_model}.
Thus, we construct $\mathcal{N}_i$ so that it contains both negatives for video $x_i$ \emph{and} narration $y_i$.
In Section~\ref{sec:exp}, we describe how $\mathcal{N}_i$ is obtained as well as evaluate the benefit of this symmetric approach.

\section{Experiments}
\label{sec:exp}
We first describe implementation details of our method in Section~\ref{subsec:impl}. 
The datasets used in our evaluation are outlined in Section~\ref{subsec:tasks}.
We present an ablation study emphasizing key ingredients of our approach in Section~\ref{subsec:ablation}.
Finally, we compare our learnt representations to previous self and fully-supervised methods in Section~\ref{subsec:benchmark}.

\renewcommand\arraystretch{1.1}
\setlength{\tabcolsep}{3pt}
\begin{table}[t]
\resizebox{0.52\linewidth}{!}{
\begin{tabular}{l|c}
\hline
Operation & output size \\
\hline
Input video & 32$\times$200$\times$200$\times$3 \\
I3D / S3D $\rightarrow$ Mixed\_5c & 4$\times$6$\times$6$\times$1024 \\
Global avg pool & 1$\times$1$\times$1$\times$1024 \\
Linear & 1$\times$1$\times$1$\times$512 \\
\hline
\end{tabular}
}
\resizebox{0.41\linewidth}{!}{
\begin{tabular}{l|c}
\hline
Operation & output size \\
\hline
Embedding & 16$\times$300 \\
Linear + ReLU & 16$\times$2048 \\
Max pool & 1$\times$2048 \\
Linear & 1$\times$512 \\
\hline
\end{tabular}
}

\caption{\small{Video (left) and text (right) model architectures.}}
\label{tab:video_text_architecture}
\end{table}

\begin{table*}
	\begin{subtable}[t]{.32\linewidth}
		\tablestyle{2pt}{1.05}
		{
			%\vspace*{1.25mm}		
			\caption{\textbf{Training loss}} \label{tab:ablation-loss}
			\vspace*{1.25mm}\vspace*{1.25mm}\vspace*{0.95mm}
			\begin{tabular}[t]{@{}l|ccccc@{}}
				Loss & \rYC & \rMSRVTT & \rCrossTask & HMDB & UCF \\
				\hline
			    Binary-Classif & 18.5 & 23.1 & 32.6 & 44.2 & 68.5\\
				Max margin & 16.3 & 24.1 & 29.3 & \textbf{56.2} & 76.6\\
				NCE & \textbf{29.1} & \textbf{27.0} & \textbf{35.6} & 55.4 & \textbf{77.5}\\
		\end{tabular}}
	\end{subtable}
	\begin{subtable}[t]{.33\linewidth}\centering
		\tablestyle{2pt}{1.05}
		{ 		
			\caption{\textbf{Negatives per positive}}\label{tab:ablation-negatives}
			\vspace*{1.25mm}\vspace*{1.25mm}\vspace*{0.95mm}
			\begin{tabular}[t]{l|ccccccc}
				$\|\mathcal{N}\|$& \rYC  & \rMSRVTT  & \rCrossTask      & HMDB & UCF     \\ \hline
				64   & 26.0 & 25.5 & 33.1 & 56.1 & 76.0  \\
				128 & 27.1 & 26.4 & 33.3  & \textbf{57.2} & 76.2 \\
				256 & 28.7 & 28.7 & \textbf{36.5}  & 56.5 & \textbf{77.5} \\
				512 & \textbf{28.8} & \textbf{29.0} & 35.6  & 55.4 & 77.4 \\ 
			\end{tabular}
		}%\vspace*{2.25mm}			
	\end{subtable}
	\begin{subtable}[t]{.33\linewidth}
		\tablestyle{2pt}{1.05}
		{ 	
			\caption{\textbf{Number of positive candidate pair}} \label{tab:ablation-positives}
			\begin{tabular}[t]{l|cccccc}
				&   NCE  & \multicolumn{5}{|c}{MIL-NCE}  \\ 
				$\|\mathcal{P}\| \rightarrow$ & 1    & \multicolumn{1}{|c}{3}    & 5 & 9 & 17   & 33   \\ \hline
				\rYC       & 29.1 & 33.6 & \textbf{35.0} & 33.1 & 32.4 & 28.3 \\
				\rMSRVTT       & 27.0 & 30.2 & \textbf{31.8} & 30.5 & 29.2 & 30.4 \\
				\rCrossTask      & 35.6 & \textbf{37.3} & 34.2 & 31.8 & 25.0 & 25.0 \\
				HMDB       & 55.4 & \textbf{57.8} & 56.7 & 55.7 & 54.8 & 51.4 \\
				UCF       & 77.5 &  79.7 & \textbf{80.4} & 79.5 & 78.5 & 77.9\\
			\end{tabular}
		}	 \vspace{-1mm}		
	\end{subtable}
	\\
	\begin{subtable}[t]{.32\linewidth}\centering
		\tablestyle{2pt}{1.05}
		%\vspace*{1mm}
		{ 
			\caption{\textbf{MIL strategy}} \label{tab:ablation-mil}
			\begin{tabular}[t]{@{}l|ccccc@{}}
				%				\toprule
				Method & \rYC & \rMSRVTT & \rCrossTask  & HMDB & UCF \\
				\hline
				Cat+NCE  & 31.9 & 30.8 & \textbf{35.2} & 56.3 & 78.9 \\
				Max+NCE  & 32.3 & 31.3 &  32.2 & 55.3 & 79.2 \\
				Attn+NCE & 32.4 & 30.2 & 33.4 & 55.2 & 78.4 \\
				%				\midrule
				MIL-NCE & \textbf{35.0} & \textbf{31.8} & 34.2 & \textbf{56.7} & \textbf{80.4} \\
				%				\bottomrule
		\end{tabular}}
	\end{subtable}%
	\begin{subtable}[t]{.34\linewidth}
		\tablestyle{2pt}{1.05}
		%\vspace*{1mm}
		\caption{\textbf{Symmetric vs asymmetric negatives}}  \label{tab:ablation-symmetric}
		{   	\begin{tabular}[t]{@{}l|ccccc@{}}
				%				\toprule
				Negatives & \rYC & \rMSRVTT & \rCrossTask & HMDB  & UCF  \\
				\hline
				$(x|y)$ & 34.4 & 29.0 &  33.9 & 55.1 & 78.1  \\
				$(y|x)$ & 19.3 & 19.4 & 28.2 &  \textbf{57.1} & 79.2\\
				$(x,y)$ & \textbf{35.0} & \textbf{31.8} &  \textbf{34.2}  & 56.7 &  \textbf{80.4} \\
				%				\bottomrule
		\end{tabular}}
	\end{subtable}
	\begin{subtable}[t]{.33\linewidth}
		\tablestyle{2pt}{1.05}
		{		\caption{\textbf{Language models}} \label{tab:ablation-language}
			%			\vspace*{1mm}
			\begin{tabular}[t]{@{}l|ccccc@{}}
				Text model & \rYC & \rMSRVTT & \rCrossTask & HMDB & UCF  \\
				\hline
				LSTM & 16.6 & 15.6 & 23.8 & 53.1 & 80.1 \\
				GRU & 16.8 & 16.9 & 22.2& 54.7 & \textbf{82.8} \\
				Transformer & 26.7 & 26.5 & 32.7 & 53.4 & 78.4 \\
				NetVLAD & 33.4 & 29.2 & \textbf{35.5} & 51.8 & 79.3 \\
				Ours & \textbf{35.0} & \textbf{31.8}  & 34.2 &  \textbf{56.7} &  80.4 \\
			\end{tabular}%\vspace*{3.5mm}`
			
		}		
	\end{subtable}
	\vspace{-1mm}	
	\caption{Ablation studies}\label{tab:1}
	\vspace{-1mm}	
\end{table*}

\subsection{Implementation details}
\label{subsec:impl}

\noindent
\textbf{Model and Inputs.}
For the 3D CNN backbone, we use the standard I3D implementation from~\cite{carreira2017quovadis} for all ablation studies and for the comparison to state-of-the-art, we report result on both I3D and S3D~\cite{xie2018rethinking}.
We use the Google News self-supervised pre-trained word2vec (d=300) embedding from~\cite{mikolov13efficient} for our word representation.
Each video clip at training contains 32 frames sampled at 10 fps (3.2 seconds) with a 200x200 resolution (224x224 at test time). 
For each narration, we take a maximum of 16 words.
More details about the model architecture and input dimensions are provided in Table~\ref{tab:video_text_architecture}.
A detailed illustration of the architecture is also given in Appendix (section~\ref{model-architecture}).

\noindent
\textbf{Visual representations evaluation.}
We evaluate our visual representations at two different semantic levels.
First, we use the output of the I3D (or S3D) \emph{Global avg pool} (see Table~\ref{tab:video_text_architecture}), to evaluate our representation for action recognition, action segmentation and action localization.
Next, the output of the last I3D (or S3D) \emph{Linear} layer (see Table~\ref{tab:video_text_architecture}), which maps the video to the joint text-video semantic space, is used in conjunction with the output of the language model for the text-video retrieval tasks.

\noindent
\textbf{Training dataset.}
We train our model using the HowTo100M~\cite{miech19howto100m} narrated video dataset.
It consists of more than 1.2M videos accompanied with automatically generated speech transcription.
We use the provided transcription to create pairs of video / caption defined by each caption time stamp.
Note that while the original dataset~\cite{miech19howto100m} consists of 136M pairs, we only used 120M pairs to comply with the YouTube wipe out policy.
Each video shorter than 5 seconds is extended symmetrically in time so that the duration is at least 5 seconds.
Then we randomly sample, a fixed length clip of 3.2 seconds within each video at training.
For each clip-narration training pair $(x,y)$ sampled, we construct the bag of positive candidate pairs $\mathcal{P}$ by considering the nearest captions in time to $y$ as depicted in Figure~\ref{fig:mil_vizu}.
For example, if we set the number of positive candidate pairs to 3, we would have $\mathcal{P} = \{(x,y), (x,y^{1}),(x,y^{2})\}$ where $y^{1}$ and $y^{2}$ are the 2 closest narrations in time to $y$.
We work with batch containing $B$ positive video-narration pairs $\{(x_{i}, y_{i})\}_{i \in [1, B]}$ .
We construct the set $\mathcal{N}$ by simply creating negative pairs from this batch by combining $\{(x_{i}, y_{j})\}_{i \neq j}$.
Since all representations are already computed, computing negative scores is cheap and efficient.

\noindent
\textbf{Optimization.}
We use the ADAM~\cite{kingma15adam} optimizer with an initial learning rate of $10^{-3}$ with linear warm up of 5k steps.
The learning rate is decayed twice by a factor of 10.
We train our model using Cloud TPUs v3~\footnote{\url{https://cloud.google.com/tpu/}}, each Cloud TPU having a batch size of 128 videos.
Given the high computational load required for training on HowTo100M, we run ablation studies on 4 Cloud TPUs and train our model for 500k steps ($\sim$ 3 days). 
For our final evaluation in Section~\ref{subsec:benchmark}, we pick the best parameters based on our ablation study and then use 64 Cloud TPUs for 400k steps (also $\sim$ 3 days) as we observed that training on bigger batch size, and thus more epochs, had a positive impact on performance.

\subsection{Downstream tasks}
\label{subsec:tasks}
To show the generality of our learnt representations, 
we perform evaluation on five diverse downstream tasks using eight datasets described below.

\noindent
\textbf{Action Recognition:} \textit{HMDB-51~\cite{kuehne2011hmdb}, UCF-101~\cite{soomro2012}, Kinetics-700~\cite{carreira2019short}}.
We evaluate our video-only representation on the traditional HMDB-51 / UCF-101 as well as the recent Kinetics-700 action recognition tasks.

\noindent
\textbf{Text-to-Video retrieval:} \textit{YouCook2~\cite{youcook2}, MSR-VTT~\cite{xu16msrvtt}}. We use the YouCook2 and MSR-VTT text-to-video retrieval benchmarks to evaluate our off-the-shelf learnt joint text-video representation.
We follow the same evaluation protocol as described in~\cite{miech19howto100m}.
We report the retrieval performance using the recall at K (R@K) metric (with K=1,5,10) which measures the percentage of clips retrieved at the top K (the higher the better).
We also report the median rank (MedR) of videos to be retrieved (the lower the better).
Note from~\cite{miech19howto100m} that there is no intersection between YouCook2 testing and HowTo100M training videos.

\noindent
\textbf{Action Localization:} \textit{YouTube-8M~\cite{youtube8m} Segments.} 
We evaluate our video representation on YouTube-8M Segments\footnote{\url{https://research.google.com/youtube8m}}, a subset of the YouTube-8M~\cite{youtube8m} with precise temporal annotation.
We follow the YouTube-8M Segments challenge evaluation protocol and report the mAP metric.\footnote{\url{https://www.kaggle.com/c/youtube8m-2019/overview/evaluation}}

\noindent
\textbf{Action Step Localization:} \textit{CrossTask~\cite{zhukov2019crosstask}.} We use the recently released CrossTask instructional video dataset
to evaluate our off-the-shelf learnt joint text-video representation on the task of action step localization. 
We perform the same evaluation protocol as in~\cite{zhukov2019crosstask} and report the average recall (\rCrossTask) metric for the localization task.

\noindent
\textbf{Action Segmentation:} \textit{COIN~\cite{tang2019coin}.} 
We evaluate our video-only representation on the COIN action segmentation task and follow the evaluation protocol of~\cite{sun2019contrastive} by reporting the frame-wise accuracy (\rCOIN).

\subsection{Ablation studies}
\label{subsec:ablation}
We perform the ablation studies on the following downstream tasks: MSR-VTT R@10 (\rMSRVTT), YouCook2 R@10 (\rYC), HMDB-51 and UCF-101 recognition accuracy on split 1 and CrossTask average recall (\rCrossTask). % as they are the less computational evaluation tasks.
This subset of downstream tasks has been chosen for their simplicity of evaluation and because they cover a wide range of tasks.

\noindent
\textbf{Which loss is better for learning the joint embedding?}\\
In this ablation study (Table~\ref{tab:ablation-loss}), we compare different losses for matching the text and video embeddings in the
standard single-instance learning setting where we pair each video clip to its closest narration in time.
We compare the NCE based approach (ours) to the frequently used max margin ranking loss~\cite{chowdhury2018webly,dong19dual,hendricks17localizing,karpathy14deepfragment,miech18learning,mithun2018learning,wang2018learning,wang2016learning,wray2019fine,wu2017sampling} and a binary classification loss (\ie sigmoid cross entropy loss) that has shown to be effective in video-audio matching~\cite{arandjelovic17look,arandjelovic2018objects}.
Overall, the NCE loss outperforms other losses or works similarly on all five tested datasets.

\noindent
\textbf{The more negatives, the better.}
We keep the same single-instance learning setting and assess the quality of our representations trained with different number of sampled
negative examples per positive pair in Table~\ref{tab:ablation-negatives}.
We can see that the overall performance increases with the number of negatives.
For the rest of the ablation studies, we use 512 negative samples per positive.

\noindent
\textbf{How many positive candidates pairs to consider?}
We evaluate the benefit of going from the single-instance learning approach to the proposed multiple-instance based approach in 
Table~\ref{tab:ablation-positives}.
In this experiment, we vary the number of positive candidate training pairs $\|\mathcal{P}\|$ for each video clip from 1 (\ie single-instance learning setting) up to 33 candidates.
Adding candidates significantly improves the performance upon the single-instance learning baseline.
Moreover, we observe a trade-off between having too many candidates and not having enough of them, as we reach the best results by considering 3 to 5 positive candidates. 
We believe that adding too many contextual narrations increases the chance for irrelevant ones as they are sampled further in time from the considered video clip.
For the rest of the paper we fix the number of positive candidate pairs to 5.

\begin{figure}[t]
      \centering
		\includegraphics[width=\columnwidth]{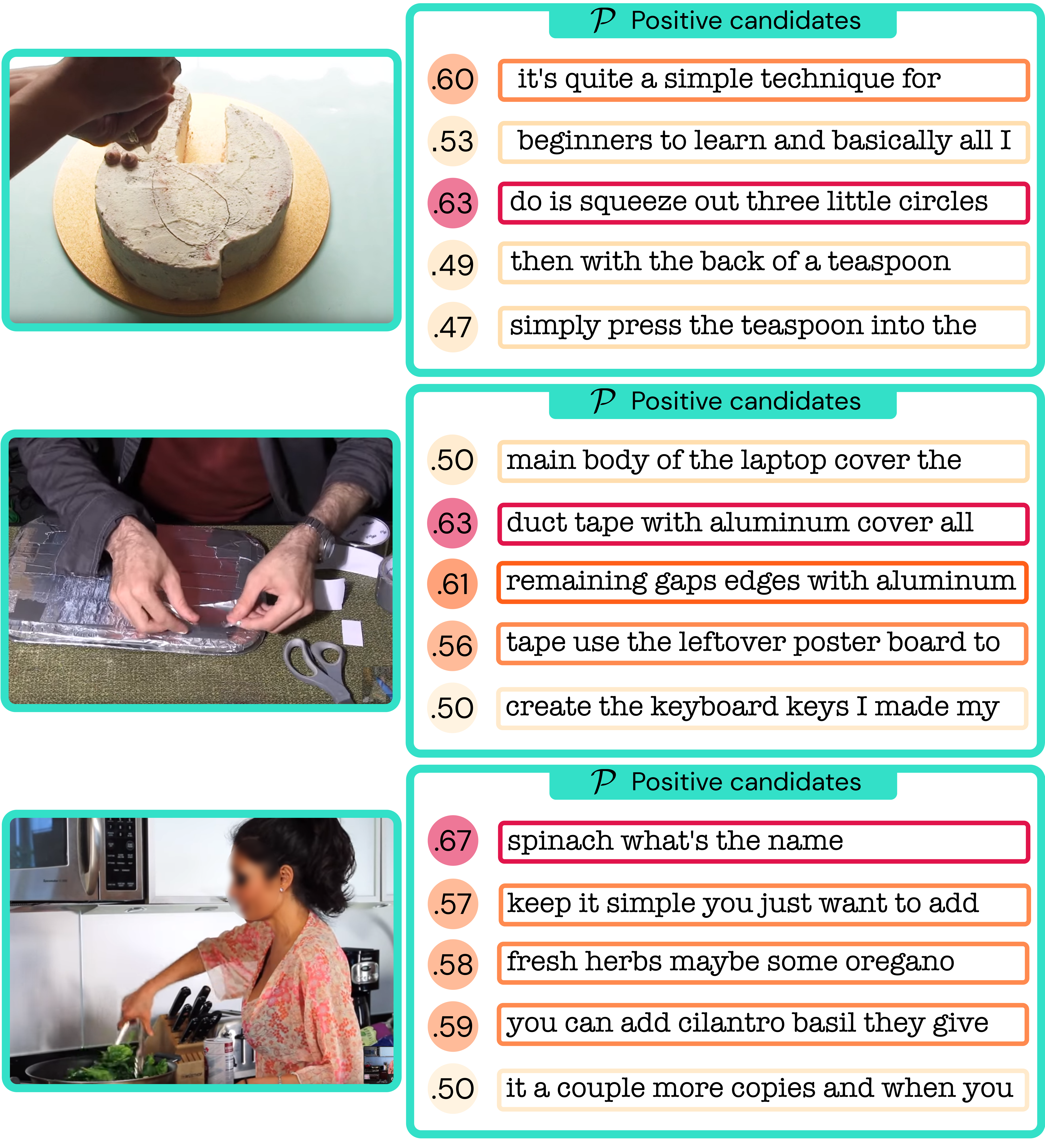}\vspace{-0.5em}
	\caption{\label{fig:qualitative} \small Selected video and narration pairs from five positive candidates on HowTo100M held-out samples using MIL-NCE.
}
\end{figure}

\noindent
\textbf{MIL-NCE vs other MIL based approaches.}
In Table~\ref{tab:ablation-mil}, we compare our MIL-NCE approach with methods that can also handle multiple possible candidate captions at training time.
The max-pool based approach~\cite{andrews2003support,arandjelovic2018objects,Oquab_2015_CVPR} (Max+NCE) only optimizes over the clip-caption pair with the highest similarity score among the positive candidates.
On the other hand, the attention-based approach~\cite{ilse2018attention} (Attn+NCE) computes cross-modal attention weights between all the clip-caption pairs and perform a weighted average of the similarity scores in order to consider the most relevant positive candidate pairs. 
More details about these baselines are provided in the Appendix (section~\ref{mil-baselines}).
Finally, we also compare to the single-instance learning baseline where we concatenate all of the candidate narrations as one longer narration (Cat+NCE).
Our proposed method outperforms these two standard approaches on five out of six tasks.
Figure~\ref{fig:qualitative} illustrates examples of selected pairs from a hold-out set of HowTo100M videos, using MIL-NCE.

\begin{table}[t]
  \setlength{\tabcolsep}{3pt}
    \centering  
        \resizebox{\linewidth}{!}{
      \begin{tabular}{@{}llclccc@{}}
      \toprule
      Method & Dataset & MM & Model & Frozen & HMDB & UCF \\
      \midrule
      OPN~\cite{lee2017unsupervised} & UCF & \xmark  & VGG & \xmark & 23.8 & 59.6  \\
      Shuffle \& Learn~\cite{misra2016shuffle}* & K600 &\xmark  & S3D & \xmark & 35.8 & 68.7  \\
      Wang \etal~\cite{wang2019self} & K400  & Flow & C3D & \xmark  & 33.4 & 61.2  \\
      CMC~\cite{tian2019contrastive} & UCF & Flow  & CaffeNet & \xmark & 26.7 & 59.1  \\
      Geometry \cite{gan2018geometry} & FC & Flow  & FlowNet & \xmark  & 23.3 & 55.1 \\
      Fernando \etal~\cite{fernando2017self} & UCF & \xmark & AlexNet & \xmark  & 32.5 & 60.3  \\
      ClipOrder~\cite{xu2019self} & UCF & \xmark & R(2+1)D & \xmark  & 30.9 & 72.4  \\
      3DRotNet~\cite{jing2018self}* & K600& \xmark & S3D & \xmark & 40.0 & 75.3  \\
      DPC~\cite{han2019video} & K400 & \xmark & 3D-R34 & \xmark  & 35.7 & 75.7  \\
      3D ST-puzzle~\cite{kim2019self} & K400 & \xmark & 3D-R18 & \xmark & 33.7 & 65.8  \\
      CBT~\cite{sun2019contrastive} & K600 &\xmark  & S3D & \cmark  & 29.5 & 54.0  \\
      CBT~\cite{sun2019contrastive} & K600 & \xmark  & S3D & \xmark  & 44.6  & 79.5 \\
      AVTS~\cite{korbar2018cooperative} & K600 & Audio & I3D & \xmark  & 53.0  & 83.7 \\
      AVTS~\cite{korbar2018cooperative} & Audioset & Audio & MC3 & \xmark  & \textbf{61.6}  & 89.0 \\
      \midrule
      \multirow{4}{*}{Ours} & \multirow{4}{*}{HTM} &  \multirow{4}{*}{Text} &  \multirow{2}{*}{I3D} & \cmark   & \textbf{54.8} & \textbf{83.4} \\
      & & &  &  \xmark   & 59.2 & \textbf{89.1} \\  
      & &  &  \multirow{2}{*}{S3D} & \cmark   & \textbf{53.1} & \textbf{82.7} \\
      & & &  &  \xmark   & 61.0 & \textbf{91.3} \\  
      \midrule
       \multicolumn{3}{c}{\textcolor{mygray}{Fully-supervised SOTA~\cite{xie2018rethinking}}} & \textcolor{mygray}{S3D} & \xmark &  \textcolor{mygray}{75.9} & \textcolor{mygray}{96.8} \\
      \bottomrule
    \end{tabular}
    }
        \caption{\small \textbf{Comparison to self-supervised methods on HMDB/UCF}. Results are reported by averaging the accuracy over the 3 splits for both datasets. *Shuffle \& Learn and 3DRotNet reported numbers are reimplemented in~\cite{sun2019contrastive} by using a better backbone (S3D). The \textit{MM} column indicates whether or not other modalities than the video frames have been used for the learning of the visual features. FC: FlyingChairs.}
      \label{table:hmdb-results}
\end{table}

\noindent
\textbf{Symmetric or asymmetric negative sampling?}
Recall that given a pair of video/narration $(x, y)$, we create $\mathcal{N}$ in a \emph{symmetric} manner by sampling negative narrations for the video $x$ \emph{and} negative videos for the narration $y$.
Table~\ref{tab:ablation-symmetric} compares that approach $(x,y)$ to \emph{asymmetric} alternatives: (i) by fixing the video $x$ and \emph{only} sampling negative captions $(y|x)$ and (ii) by fixing the narration $y$ and \emph{only} sampling negative videos $(x|y)$.
Overall, the best results are achieved when sampling \emph{jointly} the negatives $(x,y)$, \ie when we equally sample both video and narration negatives.

\begin{table*}
	\begin{subtable}[t]{.22\linewidth}\centering
		\tablestyle{2pt}{1.05}
		{      
\begin{tabular}[t]{lllc|c}
	\multirow{2}{*}{Method} & \multirow{2}{*}{Net} & \multicolumn{2}{c|}{Pretraining} & \multicolumn{1}{l}{\multirow{2}{*}{\rCOIN}} \\
	&                           & Dataset          & Labels      & \multicolumn{1}{l}{}                        \\ \hline
	\multirow{3}{*}{Ours}   & \ResNetFifty                      & ImNet            & \cmark       & 52.0                                        \\ % 9454534
	& I3D                       & K400             & \cmark       & 52.9                                      \\
	& I3D                       & K700             & \cmark       & 54.2                                        \\
	CBT \cite{sun2019contrastive}                     & S3D                       & \KinSix+\HowToM      & \cmark       & 53.9                                        \\  \hline
	Ours                    & I3D                       & \HowToM           & \xmark       & \textbf{59.4}     \\  
	Ours                    & S3D                       & \HowToM           & \xmark       & \textbf{61.0}                                      
\end{tabular}%
}
		\caption{\textbf{COIN}}\label{tab:COIN}
	\end{subtable}
\hspace{\fill}
	\begin{subtable}[t]{.12\linewidth}
		\tablestyle{2pt}{1.05}
		{ 		     
\begin{tabular}[t]{lcc|c}
	\multirow{2}{*}{Net} & \multicolumn{2}{c|}{Pretraining}  & \multirow{2}{*}{mAP} \\
	& Dataset           & Labels                             &                      \\ \hline
	I3D                       & \KinFour      & \cmark           &  73.7                 \\
	I3D                       & \KinSeven      & \cmark           &  74.0                 \\
	\ResNetFifty                 & \ImageNet          & \cmark          &  75.0                 \\ \hline
	I3D                       & \HowToM         & \xmark           &  \textbf{77.1}                
\end{tabular}\vspace*{3.5mm}		
	}	
		\caption{\textbf{YT8M-S}}\label{tab:YT8M}
	\end{subtable}	
\hspace{\fill}
	\begin{subtable}[t]{.12\linewidth}\centering
	\tablestyle{2pt}{1.05}
	{      
		\begin{tabular}[t]{lc|cc}
			\multirow{2}{*}{Init}& Net & \multicolumn{2}{c}{Top1} \\
			& & val         & test       \\ \hline
			Scratch   & I3D             & 57.0        & 55.4        \\ % 9400674
			\ImageNet    & I3D           & 59.9        & 58.2         \\ \hline % 9414804
			Ours        & I3D           & \textbf{61.1}        & \textbf{59.6}      \\ % 9393344
		\end{tabular}\vspace*{3.5mm}\vspace*{3.5mm}
}
	\caption{\textbf{\KinSeven}}\label{tab:Kinetics}
\end{subtable}
\hspace{\fill}
	\begin{subtable}[t]{.26\linewidth}\centering\vspace*{1mm}
		\tablestyle{2pt}{1.05}
		{ \begin{tabular}[t]{@{}lr|c@{}}
				Method & Labels used &  \rCrossTask \\
				\hline
				Alayrac \etal\cite{alayrac16unsupervised}  & \ImageNet +\KinFour & 13.3   \\
				CrossTask \cite{zhukov2019crosstask}   & \ImageNet +\KinFour & 22.4   \\
				CrossTask \cite{zhukov2019crosstask}  & \ImageNet +\KinFour +\CrossTask & 31.6   \\
				Miech \etal \cite{miech19howto100m}  & \ImageNet +\KinFour & 33.6   \\ \hline
				Ours (I3D) &  \textbf{None} & \textbf{36.4} \\
				Ours (S3D) &  \textbf{None} & \textbf{40.5} \\
		\end{tabular}}
		\caption{\textbf{CrossTask} (\CrossTask)}\label{tab:CrossTask}
	\end{subtable}
		\vspace*{-3.5mm}
	\caption{ \small Evaluation on action segmentation (a), localization (b, d) and recognition (c) benchmarks. \KinFour: Kinetics-400, \KinSix: Kinetics-600, \KinSeven: Kinetics-700, \HowToM: HowTo100M, \ImageNet: ImageNet, YT8M-S: YouTube-8M Segments, R50: 2D ResNet-50.}\label{tab:downstream_tasks}
	\vspace*{-1.5mm}
\end{table*}

\begin{table}
\begin{subtable}[t]{\linewidth}\centering
\tablestyle{2pt}{1.05}
{ 
    \resizebox{\linewidth}{!}{	
 \begin{tabular}{@{}llcccc@{}}
	Method & Labeled dataset used & R@1$\uparrow$ & R@5$\uparrow$ & R@10$\uparrow$  & MedR$\downarrow$  \\
	\midrule
	Random & None & 0.03 & 0.15 & 0.3 & 1675 \\	HGLMM FV CCA \cite{klein15associating} & \ImageNet  \ + \KinFour    \ + YouCook2 & 4.6 & 14.3 & 21.6 & 75   \\
	Miech \etal \cite{miech19howto100m}  &  \ImageNet   \ + \KinFour  & 6.1 & 17.3 & 24.8 & 46 \\
	Miech \etal \cite{miech19howto100m}  & \ImageNet   \ + \KinFour   \ + YouCook2  & 8.2 & 24.5 & 35.3 & 24 \\
	\midrule
	Ours (I3D) & \textbf{None} & \textbf{11.4} & \textbf{30.6} & \textbf{42.0} & \textbf{16} \\
	Ours (S3D) & \textbf{None} & \textbf{15.1} & \textbf{38.0} & \textbf{51.2} & \textbf{10} \\

\end{tabular}
}}
\caption{\textbf{YouCook2}}\label{tab:YouCook2}
\end{subtable}%
\\
%\hspace*{8mm}
\begin{subtable}[t]{\linewidth}\centering
	\vspace*{2mm}
	\tablestyle{2pt}{1.05}
	{ 
		\resizebox{\linewidth}{!}{	
      \begin{tabular}{@{}llcccc@{}}
	Method & Labeled dataset used & R@1$\uparrow$ & R@5$\uparrow$ & R@10$\uparrow$  & MedR$\downarrow$   \\
	\midrule
	Random & None & 0.01 & 0.05 & 0.1 & 500 \\
	Miech \etal \cite{miech19howto100m}  & \ImageNet    \ +  \KinFour \phantom{/ YouCook2} & 7.5 & 21.2 & 29.6 & 38 \\
	\midrule
	Ours (I3D) & \textbf{None} & \textbf{9.4} & \textbf{22.2} & \textbf{30.0} &  \textbf{35} \\
	Ours (S3D) & \textbf{None} & \textbf{9.9} & \textbf{24.0} & \textbf{32.4} &  \textbf{29.5} \\

\end{tabular}			
	}}
	\caption{\textbf{MSRVTT}}\label{tab:MSRVTT}
\end{subtable}%
	\caption{ \small Zero-shot evaluation on text-to-video retrieval.}\label{tab:text2vid}
\end{table}

\noindent
\textbf{Which language model?}
Finally, we also experiment with different language models (1 layer LSTM~\cite{hochreiter97lstm} or GRU~\cite{cho11GRU}, 1 layer and 8 attention heads Transformer~\cite{vaswani2017attention} and NetVLAD with 32 clusters~\cite{arandjelovic16netvlad}) and compare them to our simple model (see Table~\ref{tab:video_text_architecture}) in Table~\ref{tab:ablation-language}.
Even though our language model is similar to simple bag-of-word approach, it performs better on average and is more consistent over the five tasks than the other models.
In particular, our model significantly outperforms the other language models on the text-to-video retrieval tasks (\rYC \ and \rMSRVTT), where language plays the most important role.
We believe that a sophisticated language understanding is not key for our learning task.
Instead, most of the time, detecting and matching the main keywords in each narration is enough.
In Appendix (section~\ref{bert-pretrain}), we replace word2vec with a pretrained BERT model and obtain worse performance. We believe this is due to the domain gap between web text corpus and ASR outputs.

\subsection{Comparison to the state-of-the-art}
\label{subsec:benchmark}

\noindent
\textbf{Video only representation.}
In Table~\ref{table:hmdb-results}, we evaluate our learnt representations on the \mbox{HMDB-51}~\cite{kuehne2011hmdb} and UCF-101~\cite{soomro2012} action recognition benchmarks by extracting averaged pooled Mixed\_5c features from the HowTo100M pretrained backbone.
More specifically, we compare to self-supervised approaches, which similarly to our work, do not make use of any annotated video or image dataset when training the visual representation.
%For AVTS~\cite{korbar2018cooperative}, we also report performance with the same I3D~\cite{carreira2017quovadis} backbone as ours.
We outperform state-of-the-art on UCF-101 and perform on par with AVTS~\cite{korbar2018cooperative} on HMDB-51.
Most importantly, our learnt representation significantly outperforms many prior approaches even without fine-tuning.
This result is significant as it demonstrates generalization of our representation to diverse sets of actions despite being trained on uncurated instructional videos.

Next, we evaluate our visual representation on COIN~\cite{tang2019coin} action segmentation task in Table~\ref{tab:COIN}.
We split videos in subsequent clips of 1.5 seconds and represent them by concatenating three features: the local representation from I3D (or S3D), the global average pooled representation across the entire video and the relative positional embedding of the video clip. 
We train a logistic regression to predict the label for each clip.
We compare our HowTo100M pretrained I3D network to an I3D fully-supervised on Kinetics-400, Kinetics-700 as well as a ResNet-50 fully supervised on ImageNet.
We also compare to the state-of-the-art approach on COIN, CBT~\cite{sun2019contrastive}, which relies on a fully supervised S3D~\cite{xie2018rethinking} trained on Kinetics-600.
Our learnt representation performs better than representations trained on Kinetics-400, Kinetics-700 or ImageNet.
Moreover, our method also significantly outperforms the state-of-the-art CBT~\cite{sun2019contrastive} despite their use of fully-supervised representation trained on Kinetics-600 and a Transformer model.
We believe this improved localization ability is due to the fact that our model was trained on narrative descriptions with temporally localized timestamps
as opposed to coarse video-level annotations.

We also report performance on the recently released YouTube-8M Segments dataset in Table~\ref{tab:YT8M}.
Since no results have been published for this benchmark yet, we only compare the classification performance using different fully-supervised representations (\ie I3D trained on Kinetics-400 / Kinetics-700 or ResNet-50 trained on ImageNet).
Here again, our learnt representation outperforms all of the fully-supervised counterparts despite the domain gap between YouTube-8M and uncurated instructional videos.

Finally, in Table~\ref{tab:Kinetics} we investigate the benefit of initializing an I3D model with our learned weights for a large-scale action recognition dataset (\ie Kinetics-700).

We compare to a randomly initialized I3D and one inflated from an Inception network pretrained on ImageNet~\cite{carreira2017quovadis}.
We obtain a $4\%$ improvement over the randomly initialized I3D and $1.4\%$ over the ImageNet pretrained I3D~\cite{carreira2017quovadis}.

\noindent
\textbf{Joint text-video representation.}
We report text-to-video retrieval results on the YouCook2 (Table~\ref{tab:YouCook2}) and MSR-VTT (Table~\ref{tab:MSRVTT}) using our 
off-the-shelf model trained on HowTo100M.
Note that our model has not seen any YouCook2 nor MSR-VTT annotated videos, hence for fair comparison on the MSR-VTT dataset we only compare to prior work~\cite{miech19howto100m} that did not finetune on MSR-VTT.
On YouCook2, our model significantly outperforms all prior work.
More specifically, it performs better than \cite{miech19howto100m} which uses visual representation trained on Kinetics-400 + ImageNet and trains the joint text-video representation on both HowTo100M and \emph{YouCook2}.
On MSR-VTT, our method performs also better than~\cite{miech19howto100m}, yet without using any manually annotated dataset.
Finally, we also evaluate our off-the-shelf model trained on HowTo100M on the CrossTask~\cite{zhukov2019crosstask} action localization benchmark in Table~\ref{tab:CrossTask}.
We perform localization via a video-to-text retrieval approach similarly to~\cite{miech19howto100m}.
Our method outperforms state-of-the-art approaches, here again, without using manual supervision.

\section{Conclusion}
We have addressed the challenging task of learning visual representations from scratch using uncurated instructional videos.
Our approach \emph{did not rely on any manually annotated video or image dataset}.
To deal with the misalignment between narrations and videos, we have introduced MIL-NCE, a multiple instance learning approach derived from noise contrastive estimation.
We have applied MIL-NCE to the uncurated HowTo100M dataset and obtained strong visual representations that outperformed self-supervised as well as fully-supervised representations on downstream tasks.
More generally, we believe MIL-NCE can be applicable in several multiple instance learning problems where representation learning is key.

\paragraph{Acknowledgements.}
We would like to thank: Relja Arandjelovi\'c, Pauline Luc, Gunnar Sigurdsson for helpful discussions.
The project was partially supported by 
Antoine Miech Google PhD fellowship, the ERC grant LEAP (No.\,336845), the CIFAR 
Learning in Machines\&Brains program, and the European Regional 
Development Fund under the project IMPACT (reg. no. 
CZ.02.1.01/0.0/0.0/15\_003/0000468).

{\small
\bibliographystyle{ieee_fullname}
\bibliography{master-biblio}
}

\clearpage
\appendix

\section*{Appendix overview}

We provide in Section~\ref{mil-baselines} technical details about the baselines introduced in \textbf{Table 2 (d - MIL strategy)} from the ablation studies.
Finally Section~\ref{model-architecture} provides a visualization of the model architecture used in our work.

%%%%%%%%% BODY TEXT
\section{Max+NCE and Attn+NCE baselines} \label{mil-baselines}
We use the same notation as in the main paper for this section.

\noindent
{\bf Max+NCE.} This baseline aims at reproducing the standard max-pool based approach often used in multiple instance learning, but here combined with the NCE loss.
Formally, this can be written as maximizing the following objective:

\begin{equation}
\label{eq:max-nce-objective}
\max_{f,g}\sum_{i=1}^n\log\left(\text{MaxNCE}_{i}\right),
\end{equation}
where: 
\begin{equation}
\label{eq:max-nce}
\text{MaxNCE}_{i} = \frac{\max\limits_{(x,y)\in\mathcal{P}_i}e^{ f(x)^\top g(y)}}{\hspace*{-1mm}\max\limits_{(x,y)\in\mathcal{P}_i}e^{ f(x)^\top g(y)}+\hspace*{-3mm}\sum\limits_{(x',y')\sim\mathcal{N}_i}\hspace*{-3mm}e^{f(x')^\top g(y')}}.
\end{equation}

Intuitively, this corresponds to choosing the \emph{best} positive candidate pair among all pairs $\mathcal{P}_i$ according to the model.

\noindent
{\bf Attn+NCE.} 
This other baseline aims at selecting best candidate pairs via a cross-modal soft-attention mechanism between the clips and narrations. 
The cross-modal attention mechanism $a$ is defined as follows:
\begin{equation}
\label{eq:attn}
a(x, y, \mathcal{P}_i) = \frac{e^{f_{a}(x)^\top g_{a}(y)}}{\sum\limits_{(x',y')\in\mathcal{P}_i}e^{f_{a}(x')^\top g_{a}(y')}},
\end{equation}
where $f_{a}$ and $g_{a}$ are two parametrized functions.
In practice $f_{a}$ and $g_{a}$ are sharing parameters with $f$ (respectively $g$) except for the last `Linear' layer (see Figure~\ref{fig:architecture}).
Given that cross-modal attention mechansim, the Attn+NCE objective is:
\begin{equation}
\label{eq:attn-nce-objective}
\max_{f,g,a}\sum_{i=1}^n\log\left(\text{AttnNCE}_{i}\right),
\end{equation}
where: 
\begin{equation}
\label{eq:attn-nce}
\text{AttnNCE}_{i} = \frac{e^{\sum\limits_{(x,y)\in\mathcal{P}_i} a(x, y, \mathcal{P}_i) \ f(x)^\top g(y)}}{\hspace*{-3mm}e^{\sum\limits_{(x,y)\in\mathcal{P}_i} \hspace*{-3mm}a(x, y, \mathcal{P}_i) \ f(x)^\top g(y)}+\hspace*{-3mm}\sum\limits_{(x',y')\sim\mathcal{N}_i}\hspace*{-3mm}e^{f(x')^\top g(y')}}.
\end{equation}
The intuition behind this approach is to allow the model to have a separate selection mechanism for the positive candidate pairs.

\section{Model architecture} \label{model-architecture}
\begin{figure}[h]
	\centering
	\includegraphics[width=\columnwidth]{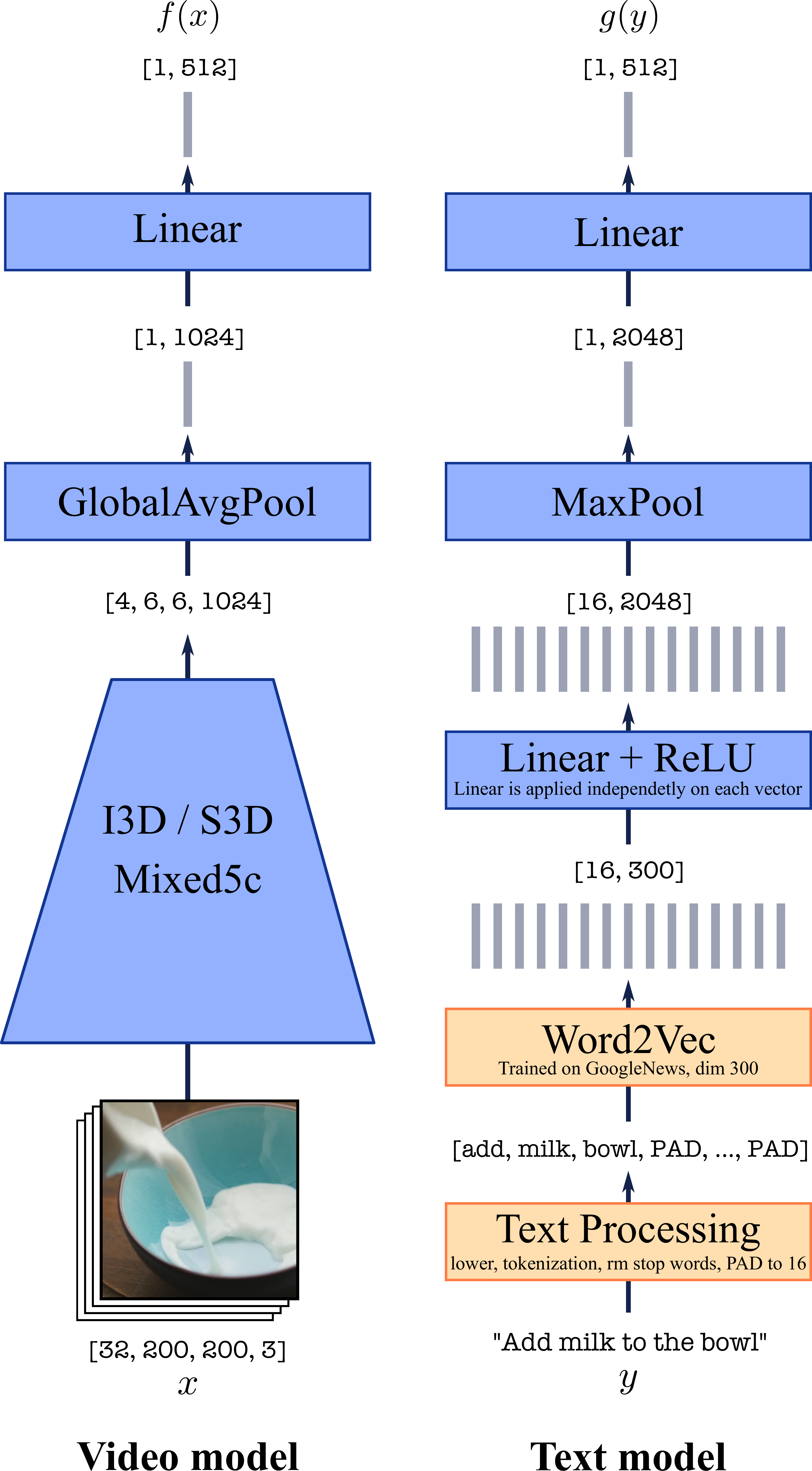}
	\caption{\label{fig:architecture} \textbf{Model architecture}. In this figure, we provide a visualization of the video embedding network $f$ (\textbf{left}) and the text embedding network $g$ (\textbf{right}). Modules displayed in blue are trained \emph{from scratch} on the challenging uncurated HowTo100M dataset using the MIL-NCE loss. The word embeddings are learned in an unsupervised fashion using Word2Vec trained on GoogleNews and are kept fixed during training.		
		Finally, the dimensions of the \emph{outputs} of each layer are specified in brackets, \eg the output of the `Word2Vec' layer is of size $[16, 300]$ corresponding to the $16$ word embedding vectors of dimension $300$ (one vector for each word, also materialized by the 16 grey rectangles).}
\end{figure}

Figure~\ref{fig:architecture} provides an illustration of the video model $f$ and text model $g$ used in the main paper.

\section{Pretrained BERT} \label{bert-pretrain}
We provide an ablation study where we replace the input text vectors coming from Word2Vec with more advanced contextual word embedding vectors from a BERT model~\cite{bertconf}.
In details, we use the BERT base model~\footnote{\url{https://tfhub.dev/google/bert_cased_L-12_H-768_A-12/1}} to replace the Word2Vec module in our model (see Figure~\ref{model-architecture}): we process the sequence of 16 input words to obtain 16 output vectors of dimension 768.
Apart from the fact that the input dimension of word vectors is increased from 300 to 768, the rest of the text model is kept the same.
For a fair comparison in terms of number of parameters, we do not finetune the BERT model.
For this experiment, we use an I3D model under the same training setting used in the ablation study~\ref{subsec:ablation}.
Also, as BERT may be sensitive to the absence of stop words, we also run an experiment where we do not remove the stop words during the text preprocessing phase.
Results are given in Table~\ref{table:bert_pretraining}.
We observe a strong degradation in performance when using BERT pretrained vectors.
We believe this is due to the domain gap between web text corpus and ASR outputs.
We leave further investigation as future work.

\begin{table}[t]
	\centering  
	\resizebox{0.7\linewidth}{!}{
	\begin{tabular}{@{}lll@{}}
	\toprule
	Input word vectors & YR10          & MR10          \\ \midrule
	BERT wo. stop words  & 19.0          & 19.7          \\
	BERT w. stop words  & 17.6          & 23.9          \\
	Word2Vec    & \textbf{35.0}  & \textbf{31.8} \\ \bottomrule
\end{tabular}
	}
	\caption{Results when using BERT vectors as inputs instead of Word2vec.}
	\label{table:bert_pretraining}
\end{table}

\end{document}